\documentclass[conference]{IEEEtran}
\IEEEoverridecommandlockouts
\usepackage{cite}
\usepackage{amsmath,amssymb,amsfonts}
\usepackage{algorithmic}
\usepackage{longtable, adjustbox}
\usepackage{graphicx}
\usepackage{textcomp}
\usepackage{xcolor}
\usepackage{float}
\usepackage{array}
\newcolumntype{L}{>{\centering\arraybackslash}m{3cm}}
\def\BibTeX{{\rm B\kern-.05em{\sc i\kern-.025em b}\kern-.08em
    T\kern-.1667em\lower.7ex\hbox{E}\kern-.125emX}}

\makeatletter
\newcommand{\linebreakand}{%
  \end{@IEEEauthorhalign}
  \hfill\mbox{}\par
  \mbox{}\hfill\begin{@IEEEauthorhalign}
}
\makeatother

\author{
  \IEEEauthorblockN{ Carlos Gómez*}
\IEEEauthorblockA{\textit{School of Computer Science} \\
\textit{Technological University Dublin}\\
Dublin, Ireland \\
carlos.g.tapia@myTUDublin.ie}
  \and
  \IEEEauthorblockN{Niamh Belton*}
\thanks{*These two authors contributed equally.}
\IEEEauthorblockA{\textit{School of Medicine} \\
\textit{University College Dublin}\\
Dublin, Ireland \\
niamh.belton@ucdconnect.ie}
  \and
  \IEEEauthorblockN{ Boi Quach}
\IEEEauthorblockA{\textit{School of Computing} \\
\textit{Dublin City University}\\
Dublin, Ireland \\
mai.quach3@mail.dcu.ie }
  \linebreakand 
  \IEEEauthorblockN{Jack Nicholls}
\IEEEauthorblockA{\textit{School of Computer Science} \\
\textit{University College Dublin}\\
Dublin, Ireland \\
jack.nicholls@ucdconnect.ie }
  \and
  \IEEEauthorblockN{ Devanshu Anand}
\IEEEauthorblockA{\textit{School of Electronic Engineering} \\
\textit{Dublin City University}\\
Dublin, Ireland \\
devanshu.anand2@mail.dcu.ie}
}

\title{A Simplistic Machine Learning Approach to Contact Tracing}
\begin{document}
\maketitle

\begin{abstract}
This report is based on the modified NIST challenge, Too Close For Too Long, provided by the SFI Centre for Machine Learning (ML-Labs). The modified challenge excludes the time calculation (too long) aspect. By handcrafting features from phone instrumental data we develop two machine learning models, a GBM and an MLP, to estimate distance between two phones. Our method is able to outperform the leading NIST challenge result by the Hong Kong University of Science and Technology (HKUST) by a significant margin.

\end{abstract}


\section{Introduction}
The outbreak of the Coronavirus disease (COVID-19) has nearly
paralyzed the global economy and caused mass disruptions
in people’s daily lives. One of the methods to control the
virus spread while optimally maintaining the normal social
operation is contact tracing. Contact tracing is the process of
identifying who has come into close contact with an infected
case so that those suspected cases can be notified and
possibly quarantined. As economies open up, digital contact tracing is emerging as an important tool to help contain the spread of COVID-19 by providing exposure notification to susceptible individuals who came in close contact to infected individuals. There have
been several proposals varying across different modalities,
but Bluetooth is the most widely emerging technology for
digital contact tracing due to the technology’s aptness for
the task. Current approaches for contact tracing rely on using Bluetooth Low Energy (BLE) signals (or chirps) from smartphones to detect if a person has been too close for too long (TC4TL) to an infected individual. However, the received signal strength indicator (RSSI) value of Bluetooth chirps sent between phones is a very noisy estimator of the actual distance between the phones and can be dramatically affected in real-world conditions by i) where the phones are carried, ii) body positions, iii) physical barriers, and iv) multi-path environments. To better characterize the effectiveness of range and time estimation using the BLE signal, many research organizations around the world are collecting Bluetooth chirp data as well as other phone sensor data (e.g., accelerometer and gyroscope) between various types of phones with simulated real-world variability.

\indent One of the key questions that is being answered in this report is
whether machine learning can aid in the accuracy of calculating distance 
between two phones. By entering a challenge and building a model on the supplied data we aim to improve the accuracy and compare it against the NIST competition submissions.

\section{Challenge Description \& Evaluation}
The National Institute of Standards and Technology (NIST), in collaboration with the MIT Private Automated Contact Tracing (PACT) research group announced a Too Close For Too Long, or TC4TL, challenge. SFI Centre for Machine Learning (ML-LABS) challenge focuses on the distance (Too Close) element only, while the time element (Too Long) is not part of the challenge. The objective of NIST's challenge is 1) to explore promising new ideas in TC4TL detection using BLE signals, 2) to support the development of advanced technologies incorporating these ideas, and 3) to measure performance of the state-of-the-art TC4TL detectors.

The data sets are divided into two subsets, fine-grain and coarse-grain distances. In the fine-grain subset the metric is calculated at three distances (1.2m, 1.8m, 3.0m). The coarse-grain subset is evaluated at a single distance (1.8m)\cite{b5}.

The challenge submissions are evaluated \cite{b5} locally by submitting the outputs of the models through the scoring software provided by NIST. Performance of the model is measured using probability of miss and the probability of false alarm.

\begin{equation} \label{Pmiss}
{\mathbf{\textbf{Pmiss}} ={\frac{Num\ of\ ref = TC4TL\ and\ hyp = not - TC4TL}{ref = TC4TL}}}
\end{equation}

\begin{equation} \label{Pfa}
{\mathbf{\textbf{Pfa}} ={\frac{Num\ of\ ref = not-TC4TL\ and\ hyp = TC4TL}{ref = not-TC4TL}}}
\end{equation}

These probabilities are then combined using a normalised decision cost function (nDCF)

\begin{equation} \label{nDCF}
{\mathbf{\textbf{nDCF}} ={\frac{w_{miss} P_{miss} +w_{fa} P_{fa}}{min(w_{miss} , w_{fa})}}}
\end{equation}

where $w_{miss}$ and $w_{fa}$ are costs associated with missed and spurious detections. NIST has set the challenge weights to be $w_{miss}$ = 1 and $w_{fa}$ = 1.

\section{Related Work}
The top model presented at the NIST competition was performed by the Hong Kong University of Science and Technology (HKUST) under the submission: "Contact-Tracing-Project" \cite{b4}. The data used in their model was Bluetooth signal, Inertial Measurement Unit (IMU), and phone transmission power (TxPower) information. Using a Deep Neural Network (DNN) Classification model, HKUST was able to produce the most accurate results of calculating the distance between two phones.

A group named 'LCD' used both Bluetooth and signal data to predict the distance between two phones \cite{b1}. They employed one-hot encoding for categorical variables and formulated the challenge as a regression problem using a Random Forest Regressor. They achieved a low error rate and placed second on the leader board.

\section{Data Resources and Description}

\subsection{Data Resources and Description}\label{data_src_desc}
The data used in the training of the models was provided by NIST through the challenge website portal. Additional data is available to competitors through the PACT GitHub page, also provided on the NIST website. NIST also produced the software scoring system, which enabled input of results and outputting a cost function score on the evaluation set.

The NIST data sets contained 15,552 train files, 8,423 test files, and 936 dev files, where each file represents the data collected between two devices over a time period. Each file is known as an event.  These are accompanied by meta data files containing the ground truth distances. The label provided is interpreted as the maximum distance that occurs in the time period. The data is collected in the form of “looks”. A “look” is a four second window where data is recorded \cite{b5}. Devices are static during looks but their distance can vary between looks. Each event has several looks. As previously outlined, an event is described as being either fine or coarse grain. There are two differences between fine and coarse grain events. The first difference is by how much the distance can vary in an event. Coarse grain events vary by up to 2.1 meters and fine grain events vary by up to 0.9 meters \cite{b5}. Secondly, the distribution of the distance labels differ from fine grain to coarse grain events. The distance label for fine grain events has a cardinality of four and the distance label for coarse grain events has a cardinality of two. The distance label distributions are shown in the table below.

\begin{table}[h!]
\begin{center}
\begin{tabular}{ c c c c c }
 Distance\_label & Fine\_Grain & Coarse\_Grain  \\ \hline
 1.2\_meters & 2563 & 0 \\ 
 1.8\_meters & 2569 & 2567  \\  
 3\_meters & 2553 & 0  \\
 4.5\_meters & 2556 & 2562 
\end{tabular}
\caption{Distance Label Distribution}
\label{table:distance_label_dist}
\end{center}
\end{table}

\subsection{Data Dictionary}\label{AA}

Table \ref{table:data_dict} displays a data dictionary explaining the contents of the training files provided by NIST for the challenge.

\begin{table}[h!]
\begin{center}
\resizebox{0.45\textwidth}{!}{
\begin{tabular}{ p{2.8cm} p{2cm} p{5cm} }
 Feature & Data Type & Description  \\ \hline
 TXDevice & Categorical & \multicolumn{1}{m{5cm}}{The device type that is transmitting the data}\\\hline
 TXPower & Categorical & \multicolumn{1}{m{5cm}}{The power value that of the transmitting device}\\\hline
 RXDevice & Categorical & \multicolumn{1}{m{5cm}}{The device type that is receiving the data}\\\hline
 TXCarry & Categorical & \multicolumn{1}{m{5cm}}{The transmitting device carriage position}\\\hline
 RXCarry & Categorical & \multicolumn{1}{m{5cm}}{The receiving device carriage position}\\\hline
 TXPose & Categorical & \multicolumn{1}{m{5cm}}{The transmitting device user pose}\\\hline
 RXPose & Categorical & \multicolumn{1}{m{5cm}}{The receiving device user pose}\\\hline
 Bluetooth & Continuous & \multicolumn{1}{m{5cm}}{The Bluetooth RSSI (Received Signal Strength Indicator) value }\\\hline
 Accelerometer & Continuous & \multicolumn{1}{m{5cm}}{Three planes (x,y,z) data provided by the internal TX accelerometer}\\\hline
 Altitude & Continuous & \multicolumn{1}{m{5cm}}{Two plane (x,y) data indicating altitude of TX device}\\\hline
 Gyroscope & Continuous & \multicolumn{1}{m{5cm}}{Three plane (x,y,z) orientation positional data of TX device}\\\hline
 Attitudes & Continuous & \multicolumn{1}{m{5cm}}{Three plane (x,y,z or pitch, roll, yaw) is the re-positioned gyroscope data with respect to Earth co-ordinates}\\\hline
 Gravity & Continuous & \multicolumn{1}{m{5cm}}{Three plane (x,y,z) gravitational acceleration data of TX device}\\\hline
 Magnetic-field & Continuous and Categorical & \multicolumn{1}{m{5cm}}{Four dimensional (x,y,z,e) data consisting of continuous x,y,z plane of magnetic fields, used for compass. The fourth value is the categorical accuracy field of [uncalibrated, low, medium, high]}\\\hline
 Heading & Continuous & \multicolumn{1}{m{5cm}}{Three plane (x,y,z) heading data of true heading, magnetic heading, and heading accuracy of TX device}
\end{tabular}}
\caption{Data Dictionary of Training File}
\label{table:data_dict}
\end{center}
\end{table}

\section{System Description}

\subsection{System Overview}
\subsubsection{ \textbf{Problem Formulation Considerations}}
This challenge allows for several ways of formulating the problem. These are outlined in this section.

1) The first approach would be to condense the information in the event files. The primary complexity in this task is standardising all events and creating variables so that they can be used in a machine learning model. This is a complicated task as each event has a varying number of looks, the amount of data recorded within each four second window varies between events and the Bluetooth and IMU data are not updated in a regular order. 

2) The second approach would be to make several distance predictions for each event, where the data would be formatted by creating a row for each new RSSI value and imputing values for all other variables. This would then be followed by performing an average/max/mean operation of the output distance for each event to get the final distance calculation.

3) The third approach would be to condense each of the looks within the event. The intuition is that it is known that the phones are static within each look, while the distance can vary from look to look. Similar to approach two, an average/max/mean operation would be performed on the output distance for each look for each event to get the final distance calculation. The limitations to this approach is that labels are not provided for each look. LCD \cite{b1} took this approach by condensing data for each look by using mean values and labelling all looks with the maximum distance label.

Approaches one and two were investigated for the purposes of this project. We implemented the first approach using features in relation to the transmitter and receiver devices, and RSSI values. This achieved state-of-the-art results. Approach two also achieved near state-of-the-art results. However, approach one was a simpler approach and resulted in higher accuracy.

As the ground truth labels for all datasets provided by NIST consist of four distinct distances, it is possible to formulate the problem as either a classification or regression task. A regression task would be a better reflection of reality and classification tasks do not account for the ordinal nature of the labels. However, classification methods perform better in this context due to the four distinct distance labels. He et al. \cite{b4} who are top of the leader board also opted for a classification model.

Both supervised and unsupervised techniques were also considered. The distance between phones can vary by a considerable amount. For fine grain, an event could have a maximum distance of 1.8 meters. However, the 1.8 meters may have only occurred at one look in the data. The remaining looks may have a distance of 0.9 meters. The data being recorded for this event would then be in relation to a close contact event at both 1.2 meters and 1.8 meters. However, the label provided states that the event is not a close contact at  1.2 meters. Therefore, the data is technically unlabelled or weakly labelled and therefore, an unsupervised approach to this problem may be a potential avenue. However, since maximum distance labels are provided, it is preferable to harness this information and employ a supervised technique. 

Having considered the above points, our state-of-the-art model employs approach one and implements a supervised classification technique. In our analysis, we prove how two different types of models following this method both achieve state-of-the-art results. These models are outlined in the following two sub-sections.
\\

\subsubsection{\textbf{Neural Network - Multi-Layer Perceptron (MLP)}}

Given two phones it is possible to compute a distance estimate \textbf{d'} using eq. \ref{distance_estimate}, where RSSI is the RSSI signal measured by the receiving phone, N depends on environmental factors ranging from 2 (good conditions) to 4 (bad conditions) and measured Power (TX) indicates the 1 meter RSSI \cite{b7}.
\begin{equation} \label{distance_estimate}
{\mathbf{\textbf{d'}} =10^{\frac{{TX - RSSI}}{10*N}}}
\end{equation}
The one meter RSSI is a value that is intrinsic to the trasmitter's phone BLE beacon which specifies the expected RSSI signal measured by a receiver phone at a 1 meter distance. In order to use eq. \ref{distance_estimate} we had to set the values for TX, RSSI and N. TX and N mostly depend on external factors and so at first they were set to be the constant for every file (TX = -61.02 , N = 2.187). The RSSI value was set by calculating the mean strength of the bluetooth signals, this meant scanning all signals for every file, processing the bluetooth signal strengths and disregarding the rest of the data. With values for TX, N and RSSI we used eq. \ref{distance_estimate} to calculate a distance estimate for each file. Table \ref{table:c_m1} shows the results of evaluating the predictions made to the test set using the evaluation software. 

\begin{table}[h!]
\begin{center}
\begin{tabular}{ c c c c c }
 Subset & D & P\_miss & P\_fa & nDCF  \\ \hline
 fine\_grain & 1.20 & 0.29 & 0.45 & 0.74 \\ 
 fine\_grain & 1.80 & 0.22 & 0.62 & 0.84 \\  
 fine\_grain & 3.00 & 0.14 & 0.64 & 0.78 \\
 coarse\_grain & 1.80 & 0.25 & 0.63 & 0.88 \\ \hline
 total\_error & & 0.90 & 2.34 & \textbf{3.24}
\end{tabular}
\caption{Distance estimate formula results}
\label{table:c_m1}
\end{center}
\end{table}

\begin{figure}[H]
\centering
\includegraphics[width=0.45\textwidth]{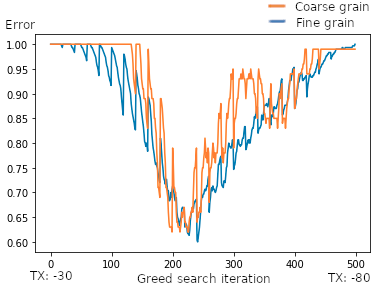}
\caption[]{Greed search for eq. \ref{distance_estimate} TX and N constants}
\label{fig:greed_search}
\end{figure}

Having access to the evaluation software it was possible to test different values TX and N and evaluate all of the results to then pick the best performing values for TX and N. A range was set for TX [-30,-80] and N [2,4]. Since every file is assigned a coarse/fine grain label two different set of parameters were obtained, one for coarse grain files and one for fine grain files. This way we can use different TX and N values for the distance estimate formula depending on a file fine/coarse grain label. The greed search results can be seen in figure \ref{fig:greed_search}. This allowed us to find the better performing TX and N constants on the training set for fine and coarse grain. CG: [TX:-52,N:2.6], FG [TX:-54,N:2.1]. 

With this approach we were able to improve the previous model results by a significant margin. Results for this method can be seen on table \ref{table:c_m2}. Note that the testing set was not used for setting the values of the constants. The values were adjusted for the training data and then applied to the testing data.

\begin{table}[h!]
\begin{center}
\begin{tabular}{ c c c c c }
 Subset & D & P\_miss & P\_fa & nDCF  \\ \hline
 fine\_grain & 1.20 & 0.56 & 0.10 & 0.66 \\ 
 fine\_grain & 1.80 & 0.47 & 0.15 & 0.62 \\  
 fine\_grain & 3.00 & 0.29 & 0.25 & 0.54 \\
 coarse\_grain & 1.80 & 0.53 & 0.11 & 0.64 \\ \hline
 total\_error & & 1.85 & 0.61 & \textbf{2.46}
\end{tabular}
\caption{Distance estimate + greed search results}
\label{table:c_m2}
\end{center}
\end{table}

Even though the improved version of the distance model yielded better results than the original model the error rate was far from being perfect. Considering that Bluetooth RSSI signals contain high quantities of noise and depend on environmental factors it is not possible to compute an accurate distance prediction using just the RSSI signal. In order to overcome this issue the improved version of the model uses other sensor information, information that we were discarding before. The predicted distance will essentially be the input to a classifier along with other available phone data in order to make the model learn to filter the error noise caused by the RSSI noise. Because of the way the evaluation software is defined there are only 4 possible labels for the predicted distance [1.2, 1.8, 3.0, 4.5]. This is the reason why we are building our model to be a classifier, this means having only 4 (num\_classes) neurons in the output layer and setting the last activation function to be the softmax function. The rest of the activation functions were set to be the ReLU (Rectified Linear Unit) function as they empirically obtained the best validation results. The architecture of the model is a vanilla neural network with 3 hidden layers with 128 neurons each. The training process had 4 epochs with batches of size 128 and the Adam optimizer was used to reduce the mse loss function which was used to compare the predictions made by the model against the train labels. Dropout layers were not used as they did not had an effect on the output or had a bad effect if the drop probability was set too high. The input to the network for every training iteration is a preprocessed version of the training files, either one file at a time or in batches. This is further detailed on table \ref{table:c_m3_input}.

\begin{table}[h!]
\begin{center}
\resizebox{0.45\textwidth}{!}{
\begin{tabular}{ p{2.8cm} p{2cm} p{5cm} }
 Input & Data & Description  \\ \hline
 Predicted distance & Continuous (1.2,4.5) & \multicolumn{1}{m{5cm}}{Distance estimate using equation \ref{distance_estimate}. Coarse grain: [TX:-52,N:2.6], Fine grain [TX:-54,N:2.1]. RSSI is the mean RSSI signal strength for the file} \\\hline
 Normalized* mean RSSI & Continuous (0,1) & \multicolumn{1}{m{5cm}}{Mean RSSI signal strength for the file}\\\hline
 Normalized* path loss attenuation &Continuous (0,1) & \multicolumn{1}{m{5cm}}{Transmit power - 41 - mean RSSI}\\\hline
 Coarse/fine grain & Categorical (0,1) & \multicolumn{1}{m{5cm}}{0 for fine grain and 1 for coarse grain}\\\hline
 TXPower & Categorical(0,2) & \multicolumn{1}{m{5cm}}{0 for TX power 7 / Unknown, 1 indicates TXPower 8 and 2 indicates TX power 12}\\\hline
 TXCarry & Categorical(0,2) & \multicolumn{1}{m{5cm}}{{0 for unknown transmitter carry state, 1 for hand and 2 for pocket}}\\\hline
 RXCarry & Categorical(0,2) & \multicolumn{1}{m{5cm}}{Same encoding as TXCarry for receiver phones}\\\hline
 TXPose & Categorical(0,2) & \multicolumn{1}{m{5cm}}{0 for unknown transmitter pose, 1 for sitting and 2 for standing}\\\hline
 RXPose & Categorical(0,2) & \multicolumn{1}{m{5cm}}{Same encoding as TXPose for receiver phones}\\\hline
 TXDevice & Categorical(0,2) & \multicolumn{1}{m{5cm}}{Model of the transmitter phone. 0 for unknown / older than iPhone 7, 1 for iPhone 7-8, 2 for iPhone X or newer}\\\hline
 RXDevice & Categorical(0,2) & \multicolumn{1}{m{5cm}}{Model of the receiver phone. Same encoding as TXDevice}\\
 
\end{tabular}}
\caption{Preprocessed file, input to the MLP\\ 
\scriptsize *Mean RSSI and path loss attenuation are min-max normalized}
\label{table:c_m3_input}
\end{center}
\end{table}

\emph{\textbf{Evaluation}}

The MLP model was evaluated using the software provided by the challenge organizers. The model was trained using the training data, 10\% of which was used for validation, using the rest for training. The trained model was evaluated for the predictions it made for the testing data having access to the test files metadata (coarse/fine grain label, phone carriage state) without access to the target labels other than for evaluating the output. Results for the MLP method can be seen on table \ref{table:c_m3}.

\begin{table}[h!]
\begin{center}
\begin{tabular}{ c c c c c }
 Subset & D & P\_miss & P\_fa & nDCF  \\ \hline
 fine\_grain & 1.20 & 0.43 & 0.19 & 0.61 \\ 
 fine\_grain & 1.80 & 0.33 & 0.14 & 0.48 \\  
 fine\_grain & 3.00 & 0.35 & 0.20 & 0.55 \\
 coarse\_grain & 1.80 & 0.30 & 0.14 & 0.44 \\ \hline
 total\_error & & 1.41 & 0.67 & \textbf{2.08}
\end{tabular}
\caption{MLP results}
\label{table:c_m3}
\end{center}
\end{table}

\subsubsection{\textbf{Tree-based Method - Gradient Boosting Machine}}
A Gradient Boosting Machine (GBM) is a sequential, tree-based, ensemble technique for supervised classification and regression tasks. A GBM was chosen as our second model as it has shown consistent superior performance on tabular data in comparison to similar algorithms such as random forest. The GBM was implemented in H2O, an open-source machine learning platform. 

\emph{{\textbf{Feature Selection}}}
\\The features included in the model are shown in table V with the exception of the normalised features and the coarse grain feature. These features are easily condensed into one row as they remain unchanged throughout an event. An advantage of tree based methods is that they automatically perform feature selection. The model is unlikely to use a feature if it not informative of the target value and therefore, feature selection methods are not required. 

\emph{\textbf{Data Pre-processing}}
\\GBMs in H2O do not require pre-processing of categorical variables prior to modelling. H2O handles categorical variables using the 'nbins\_cat' parameter. This parameter specifies the number of bins to group the levels of a variable. The default value is 1024. The maximum number of levels of a variable, in this case, is 15. Therefore, each level of each variable would have its own bin. Levels are assigned integers and the best split is then determined. The integer assigned to a level is arbitrary. For example, a node could split on a categorical variable where levels assigned an integer of zero and two are split from levels assigned an integer of one and three\cite{b2}. Therefore, it is different to label encoding. This method of categorical encoding is susceptible to overfitting when there are variables of high cardinality. However, this is not a concern, in this case, as the feature with the highest cardinality is 15. 

Another useful feature of H2O’s GBM is that it outputs a warning message when employing the model to predict on unseen data if the unseen data contains levels that the model has not trained on i.e. levels that do not exist in the training data. This highlighted that the receiver device had three levels in the test data that were not present in the training data. These were ‘iPhone 11 Pro’, ‘iPhone11 Pro Max’ and the ‘iPhone6s’. It is assumed that any information that the receiver device may have in relation to calculating the distance between phones is related to the hardware of the phone. The intuition is that the hardware may influence other data such as the RSSI values. Therefore, the value of these unseen levels were modified to be equal to devices from the training data that have similar hardware. For example, the iPhone 6s is known is have the same Bluetooth chip as the iPhone 7. The iPhone 7 is present in the training data and therefore, cases where receiver device was equal to iPhone 6s were changed to equal the iPhone 7. 

\emph{\textbf{Modelling}}
\\Two models were built, one for fine grain and another for coarse grain data. He et al. \cite{b4} also trained separate models for fine and coarse grain. The coarse grain model was a binary classification problem and the fine grain was a multi-class classification problem. In order to select the optimal parameters, the training data was split into 80\% train and 20\% validation. 

The models were tuned using grid search where the parameters ‘ntrees’, ‘max\_depth’, ‘col\_sample\_rate’ and ‘sample\_rate’ were varied. These parameters are outlined in the appendix. The grid search strategy was to train a model and assess it using every possible combination of parameters that were specified. For each combination of parameters, a model was trained on the training data and tested on the validation data. The metric to evaluate the binary classification model was Area Under the Curve (AUC) and mean per class error for multi-class classification. Early stopping rounds was used to prevent overfitting. Early stopping rounds automatically stops the training of the model if the metric calculated on the validation data has not improved for a specified number of iterations by a specified amount. The parameters of the model with the best evaluation metric on the validation data were noted and the final model was trained using these parameters on 100\% of the training data. Prior to merging the training and validation data, a confusion matrix was constructed using the validation data for the binary classification problem to maximise the f1 score. The f1 score is the harmonic mean of precision and recall. H2O automatically selects the optimal threshold to maximise the f1 score. The threshold had a value of 0.56 and it was noted in order to transform model output to distance predictions when employing the model on test data. 

\emph{\textbf{Evaluation}}
\\The most important features in the final model included the predicted distance and transmitter device. It is possible that an interaction exists between these two features as the hardware in the device may influence other data collected by the device.

The model was evaluated on the official test data from NIST. The test data was subject to the pre-processing steps outlined earlier. The optimal threshold of 0.56 was used to convert model output to a distance prediction. Values with a model output greater than 0.56 were assigned a distance of 1.8 meters and values with a model output less than 0.56 were assigned a distance of 4.5 meters. Assigning predictions was a more complex task for the multi-class classification problem. Usually, a case would be assigned the class that the model gave the highest probability to. However, as the objective of the task is to calculate the maximum distance that occurs, it would be preferable to weight the higher distance classes. A brief analysis was conducted to investigate if weighting the larger distance classes would improve model performance. The development set of 935 events was employed for this analysis. The first step of the analysis was to assign a prediction of 4.5 meters to a case if the case had a probability greater than 0.25 of belonging to the 4.5 meters class regardless of if another distance class had a larger probability. The intuition is that if the model has no information, it would assign all cases a value of 0.25. However, the model has found some information in the data to suggest that the case has a probability greater than random chance of having a distance of 4.5 meters. When testing this hypothesis on the development set, adding this step to assigning classes decreased the accuracy slightly from when there was no such step included. The threshold was increased until the accuracy was equivalent to when this step was not included in the model. The value of this threshold was 0.28 and this was used when predicting on the test data.

\section{Results and Discussion}
The results obtained with our proposed models are shown in table \ref{table:results}, along with results from current state-of-the-art methods. The table shows that both the MLP and GBM were able to significantly reduce the average error in comparison to current state-of-the-art methods. 

\begin{table} [!h]
\centering
\resizebox{0.45\textwidth}{!}{
    \begin{tabular}{p{2.4cm} | p{0.7cm} p{0.7cm}p{0.7cm}p{0.7cm}||p{1.0cm}}
    \textbf{} & Fine Grain 1.2m & Fine Grain 1.8m & Fine Grain 3.0m & Coarse Grain 1.8m & Average nDCF \\ 
    
    \hline
    GBM (proposed) & \text{0.6} & \text{0.52} & \text{0.58} & \text{0.37} & \textbf{0.5175} \\
    
    \hline
    MLP (proposed) & \text{0.61} & \text{0.48} & \text{0.55} & \text{0.44} & \textbf{0.52} \\
    
    \hline
    Contact-Tracing-Project \cite{b4} & \text{0.68} & \text{0.54} & \text{0.59} & \text{0.41} & \text{0.555} \\
    
    \hline
    LCD \cite{b1} & \text{0.6} & \text{0.58} & \text{0.63} & \text{0.55} & \text{0.59} \\
    
    \end{tabular}}
    \caption{Results}
\label{table:results}
\end{table}

Given the significant architectural differences of the MLP and GBM, these results demonstrate how a supervised classification technique  using the handcrafted features outlined in table V is a well-founded approach to the problem. In addition to achieving state-of-the-art performance, the training time for both models is less than thirty seconds on a CPU. The success of this simplistic approach can be mainly attributed to the 'predicted distance' variable which is based on received RSSI signals. Both models saw a drastic performance improvement when adding this handcrafted feature, essentially learning to filter RSSI signal noise. This feature interacted with variables relating to the transmitter and receiver devices to further improve the accuracy of the model. This highlights how a fusion of machine learning and domain knowledge can result in superior performance.

\section{Future Work}

\begin{figure}[H]
\centering
\includegraphics[width=0.45\textwidth]{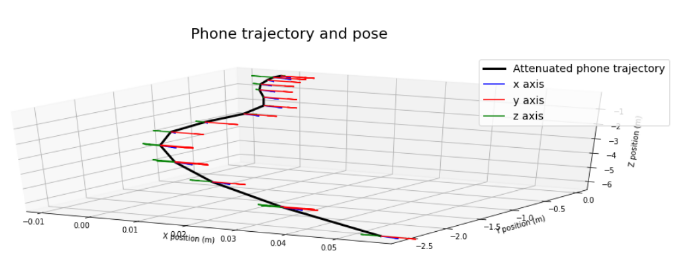}
\caption[]{Phone trajectory and pose of a random 4 seconds window}
\label{fig:trajectory}
\end{figure}

In the first approach, we use BLE RSSI is the main feature for the proximity detection since RSSI values and the distance are high correlation. However, IMU (inertial measurement unit) data which are extracted from each 4-second windows can be further decomposed to linear acceleration. By reading the explanation from the official Apple's website \cite{b6}, we implemented preprocessing to generate the linear acceleration. After that, the phone’s motions can be calculated by integrating these linear accelerations along a 4-second interval. The outputs will be a positive feature to combine with the distance values which are extracted by the formula based on the RSSI values. The negative aspect is the accuracy of MEMS accelerometer in iPhone devices since they generate a bunch of noise. If the linear acceleration is integrated, these residuals are also amplified. Thus, the challenge is finding a reasonable method which is ability of alleviating the noise. There are many methods proposed to improve the accuracy of trajectory such as Fourier Transform, sensor fusion and Kalman filtering. We made an effort to apply Fourier Transform to obtain the attenuated trajectory. The figure \ref{fig:trajectory}  is the random sample’s result within 4 seconds window. We built quiver plots to plot sets of arrows that indicated the phone’s [x,y,z] axes onto the trajectory. 

In the future, we try to use a variety of method to decide which types of filter will indicate exactly the accuracy of phone's motion.

The GBM could be further improved by investigating the following in the future. 

1) H2O offers functionality to create a custom metric to optimise the tuning of the model. Further analysis could be conducted on creating a Normalised Decision Cost Function metric for tuning and training.

2) Performing an analysis on the possibility of an interaction existing between the transmitter device and the RSSI values would be beneficial for future modelling.

3) The threshold analysis for weighting higher distance classes could be further investigated. Presently, it was investigated for the maximum distance class of 4.5 meters. However, this could be expanded to also weight the 3 meters class.

As for the proposed MLP model we left unexplored the possibility of considering other sensor data such as accelerometer or gyroscope. We believe including this extra information could help the model account for the RSSI noise and further reduce the classification error. We also left unexplored the possibility of trying different, more powerful, network architectures for the same input features. We believe this could also improve the accuracy of the model and we leave the question open for research.

\section*{Acknowledgment}
This work was funded by Science Foundation Ireland through the SFI Centre for Research Training in Machine Learning (18/CRT/6183)\\
We would also like to acknowledge Carles García Cabrera for mentoring us throughout the project.

\section{Appendix }
\subsection{Appendix A}

\begin{table}[h!]
\begin{center}
\resizebox{0.45\textwidth}{!}{
\begin{tabular}{ p{2.8cm} p{2cm} p{5cm} }
 Parameter & Description  \\ \hline
 ntrees &  \multicolumn{1}{m{5cm}}{The number of trees in a model. Default value is 50  \cite{b3}.} \\\hline
 max\_depth  & \multicolumn{1}{m{5cm}}{This is the maximum depth of a tree. Default value is 5  \cite{b3}.}\\\hline
 col\_sample\_rate  & \multicolumn{1}{m{5cm}}{The sampling rate/fraction of the columns/variables for each tree. Default value 1  \cite{b3}.}\\\hline
 sample\_rate & 
 \multicolumn{1}{m{5cm}}{The sampling rate/fraction of the rows/cases for each tree. Default value 1  \cite{b3}.}\\
 
\end{tabular}}
\caption{GBM Parameters\\ 
}
\label{table:c_m8_input}
\end{center}
\end{table}

\end{document}